\documentclass[10pt,twocolumn,letterpaper]{article}

\usepackage{cvpr}              

\usepackage{graphicx}
\usepackage{amsmath}
\usepackage{amssymb}
\usepackage{booktabs}
\usepackage[pagebackref,breaklinks,colorlinks]{hyperref}

\usepackage{rotating}
\usepackage{enumitem}
\usepackage{dcolumn}

\usepackage[capitalize]{cleveref}
\crefname{section}{Sec.}{Secs.}
\Crefname{section}{Section}{Sections}
\Crefname{table}{Table}{Tables}
\crefname{table}{Tab.}{Tabs.}

\def\confName{CVPR}
\def\confYear{2025}

\begin{document}

\title{Hidden Bias in the Machine: Stereotypes in Text-to-Image Models}

\author{Sedat Porikli$^*$ $\hspace{4cm}$ Vedat Porikli\thanks{Equal contribution}\\
Canyon Crest Academy  \\
{\tt\small \{sedatporikli, vedatporikli\}@gmail.com}
}
\vspace{-4pt}
\maketitle

\begin{abstract}

Text-to-Image (T2I) models have transformed visual content creation, producing highly realistic images from natural language prompts. However, concerns persist around their potential to replicate and magnify existing societal biases. To investigate these issues, we curated a diverse set of prompts spanning thematic categories such as occupations, traits, actions, ideologies, emotions, family roles, place descriptions, spirituality, and life events. For each of the 160 unique topics, we crafted multiple prompt variations to reflect a wide range of meanings and perspectives. Using Stable Diffusion 1.5 (UNet-based) and Flux-1 (DiT-based) models with original checkpoints, we generated over 16,000 images under consistent settings. Additionally, we collected 8,000 comparison images from Google Image Search. All outputs were filtered to exclude abstract, distorted, or nonsensical results. Our analysis reveals significant disparities in the representation of gender, race, age, somatotype, and other human-centric factors across generated images. These disparities often mirror and reinforce harmful stereotypes embedded in societal narratives. We discuss the implications of these findings and emphasize the need for more inclusive datasets and development practices to foster fairness in generative visual systems. 

\end{abstract}

\section{Introduction}
\label{sec:intro}


Recent advances in generative AI have given rise to Text-to-Image (T2I) models~\cite{Rombach2022, Ramesh2022, Chen2023pixartalpha, Betker2023, Flux, Sauer2024, Zhang2024T2I}, which have the power of realistic image generation from textual prompts. This technology is increasingly embedded into daily life with diverse and various applications~\cite{Jaiprakash2025}. It facilitates the creation of educational content and demonstrates significant artistic capabilities, aids creative expression and assisting designers in visualizing concepts and generating digital artwork~\cite{Wang2023}. In the retail sector, Text-to-Image (T2I) models enhance online shopping and digital marketing by generating customized and visually engaging content tailored to consumer preferences~\cite{Surya2020, Postma2024}. In the field of interior design, these models streamline the creative workflow by enabling  visualization of design concepts and supporting more efficient exploration of stylistic options~\cite{Chen2023}. The entertainment industry also benefits, utilizing T2I models to accelerate processes such as movie story-boarding, concept art development, and the creation of assets for video games~\cite{Li2019}. Additionally, T2I improves human-computer interaction by enabling cross-modal retrieval~\cite{Gu2018} and facilitating  accessibility for visually impaired individuals~\cite{Zhang2022}.

As AI-generated visual content becomes increasingly prevalent and harder to distinguish from real photographs, it is essential to critically examine the representations produced by text-to-image (T2I) models. Such models are trained on massive datasets like LAION~\cite{Schuhmann2021} scraped from the web. However, using such datasets introduces the risk of systematic errors, as statistical biases from incomplete sampling and inaccurate annotations are compounded in the models. Previous research has shown that T2I models often exhibit social biases that can lead to representational harms and further marginalize minority groups~\cite{Naik2023, Cho2023}. For example, a widely used model~\cite{Rombach2022} generated mostly white males for high-income occupations, while dark-skinned men were depicted as inmates and dark-skinned women as low-income workers. Users unaware of these biases may unintentionally propagate them, further exacerbating the issue. Alarmingly, AI-generated images are already being used in political campaigns~\cite{GOP2023}, potentially misleading the public before society adapts to this new technology.

Generative bias not only perpetuates stereotypical portrayals but also limits the diversity of imagery produced~\cite{Luccioni2023}. Some predict that 90\% of internet content will be AI-generated within the next few years~\cite{Europol2024}. If AI-generated images depicting amplified stereotypes contaminate the training data of future models, next-generation T2I models could become even more biased. We may soon reach a point where, unless visual content is corroborated by reliable sources, it will be safer to assume that it is AI-generated and potentially fabricated. 

Recognizing this issue, a growing body of recent research~\cite{Wan2024survey, Naik2023, Masrourisaadat2024, Chinchure2024, Cho2023, Shaw2025} has explored various dimensions of bias in T2I models. While much of the existing work has focused on biases related to gender, skin tone, and geocultural factors, there has been relatively limited exploration of biases in T2I models concerning other human-centric factors beyond the associations between certain professions and traits.

In this paper, we aim to explore additional types of bias inherent in T2I models to guide more systematic approaches to AI governance. We expand on existing attribute categories of bias in generative models and investigate the outputs of T2I models when prompts are related to actions, emotions, ideologies, life events, family structures, religion, place descriptions, and life events, analyzing 24,000 images generated for a set of 160 prompt topics\footnote{We will release a benchmark of prompts used in our study.} using three image sources: two popular T2I models (SD1.5 and Flux-1) and the Google image search engine. 


\section{Related Work}

Naik and Nushi~\cite{Naik2023} were among the first to explore social biases in T2I models, focusing on how gender, age, race, and location influence the depiction of occupations, personality traits, and everyday situations. Their study revealed significant biases in occupational representations, which could be mitigated by using more specific prompts. It also highlighted that certain personality traits were predominantly associated with specific demographics and showed that images generated from location-neutral prompts often resembled countries like the U.S. and Germany. Building on this, Luccioni \etal~\cite{Luccioni2023} examined how variations in gender and ethnicity markers in prompts affect the generated images. They found that T2I models reflected U.S. labor demographics but consistently underrepresented marginalized identities. In~\cite{Cho2023}, Cho \etal analyzed social biases across diffusion- and transformer-based models, showing that T2I models learned biases related to gender and skin tone from web image-text pairs. Masrourisaadat~\cite{Masrourisaadat2024} assessed several T2I models' performance in generating images of human faces, groups, and objects, providing valuable insights into the models' biases and limitations. Chinchure~\cite{Chinchure2024} proposed using counterfactual reasoning to quantify biases, helping identify and measure their influence on generated images. Wan \etal~\cite{Wan2024survey} surveyed prior studies on gender, skin tone, and geocultural biases in T2I models, noting that existing evaluation methods and mitigation strategies are insufficient. Research has explored ways to improve T2I models through multilingual data~\cite{Ye2024, Nguyen2024} and culturally aware interfaces~\cite{Shaw2025}, encouraging diverse interpretations, capturing a broader range of perspectives, and highlighting the impact of cultural context on generated imagery. 

Despite this growing body of research, benchmarks and evaluation methods have varied significantly, highlighting the lack of a unified framework for measuring biases. Furthermore, existing mitigation strategies remain inadequate, failing to comprehensively address biases in T2I models. In our paper, we build on previous studies and focus on several underexplored bias categories to facilitate more effective mitigation strategies.

\section{Method: Examining Inherent Biases}

\subsection{Experimental Setup}

To investigate biases in T2I models, we considered a wide spectrum of prompts covering various thematic groups: occupations, attributes and traits, actions, ideologies, emotions, family descriptions, place descriptions, depictions of spirituality, and life events. These prompts have been curated manually, representing a wide range of themes. We prepared up to 20 topics per category, resulting in 160 unique topics. For each topic, we assembled different prompts with diverse wording variations to capture a broad spectrum of meanings. We generated over 50 images per topic using random seeds, totaling 16,000 images. To ensure unbiased outputs, we avoided using negative prompts or prompt engineering. The generated images have been labeled by multiple human operators.

We utilized a UNet based model, Stable Diffusion 1.5, and a DiT based model, Flux-1, as the base T2I models, using the original checkpoints from HuggingFace without any LoRA variants. Stable Diffusion 1.5 was selected for its widespread use, open-source accessibility, and compatibility with prior studies, while newer versions introduce significant architectural and data changes. All images were generated at 512x512 resolution, following default settings with 20 sampling steps per generation. Distorted, unclear, abstract, or nonsensical images were excluded. Additionally, we collected 8,000 images from Google Image Search for comparison using the very same prompts we used for the generative models. This enabled a comparative analysis of bias label ratios across the T2I models, as well as with those observed in images retrieved from a typical Google search.

\subsection{Bias Types and Labeling}

While prior research often focuses on biases related to gender, skin tone, and geocultural factors~\cite{Wan2024survey}, there is a lack of consistency in their definitions~\cite{Blodgett2021}. Our work not only expands these bias definitions but also introduces simplified, more accessible labels, offering a clearer and more standardized approach.

\textbf{Gender Bias} in T2I models often mirrors societal gender stereotypes, where ``gender" refers to the perceived presentation and roles of individuals in generated images. We explored various forms of gender bias, including: 
\begin{itemize}[noitemsep,topsep=0pt]
    \item The tendency to generate a specific gender in response to gender-neutral prompts (e.g., ``a person"),
    \item The over- or under-representation of a specific gender in particular occupations or power dynamics,
    \item The association of certain characteristics (attributes, traits, ideologies, actions, etc.) with specific genders.
\end{itemize}
To ensure a comprehensive evaluation, we employed both positive and negative alternatives for each characteristics (e.g., attractive vs. unattractive) in text prompts. 

In labeling gender based on visual perception, we relied on observable visual cues. For simplicity, we used ``Female'', ``Male'', and ``Other''. Male cues include facial hair, broader shoulders, more angular facial features, and shorter hairstyles. Female cues involve absence of facial hair, softer facial features, narrower shoulders, and longer hairstyles. We acknowledge that other gender identities may blend traditionally male and female characteristics, or deviate from conventional norms. Visual cues vary across cultures and societies, and gender expression is highly personal, influencing how gender is perceived visually.


\textbf{Racial Bias} reinforces social stereotypes based on perceived racial attributes. For example, models may perpetuate bias by associating ``attractive" individuals with white skin and ``poor" individuals as people of color. We examined several forms of racial bias in T2I models, including: 
\begin{itemize}[noitemsep,topsep=0pt]
    \item The inclination to produce a particular race when none is specified in the prompt,
    \item The disproportionate representation of certain races in specific roles, actions, ideologies, etc. 
\end{itemize}
It is important to recognize that race is complex, fluid concept that must be approached with sensitivity due to its social implications. Race is often defined by physical characteristics such as skin color, facial features, and hair texture. We used these race labels that are common in the generated images: White, Black, East Asian, Latino, Middle Eastern, and Other, which aligns with definitions from the US Census Bureau~\cite{USCensus}. We opted not to use skin tone as a label due to the significant variation within racial groups and the challenges in creating a discrete skin tone palette.


\textbf{Age Bias} manifest in the portrayal of individuals based on perceived age-related traits. 

To evaluate the degree of age bias, we visually analyzed the depicted people and labeled the images into three broad age categories: young (up to early adolescent), adult, and senior. These categories were based on observable visual cues such as body size, muscle tone, wrinkles, graying or thinning hair, and posture changes. While this approach simplifies the annotation process, we recognize that age is a complex, multifaceted characteristic that may not always be fully represented through visual cues alone.


\textbf{Somatotype Bias} reflect societal stereotypes associated with different body sizes and shapes. To assess the extent of somatotype bias, we labeled the generated people into three categories based on their visual appearance: underweight, average, and overweight. These categories were determined using visible physical cues, such as body proportions, visible muscle tone, and overall body shape. While these common labels provide a straightforward framework for annotation, they are not without limitations. The assignment of these labels based on visual cues is inherently subjective, due to cultural differences and personal interpretations. We acknowledge that the process of categorizing somatotype involves a degree of subjectivity and may not fully capture the complexity of individual body diversity.

\textbf{Religious Bias} may manifest in the portrayal of individuals or scenes that reinforce stereotypes based on religious affiliation. Similar to other biases, these portrayals can perpetuate assumptions about what people from certain religious backgrounds look like, such as associating Christianity predominantly with Western imagery or depicting Muslims solely with religious attire like hijabs or beards.

We examined several forms of religious bias, including: 
\begin{itemize} [noitemsep,topsep=0pt]
    \item The generation of specific religious imagery when no religious identity is specified in the prompt,
    \item The over-representation or under-representation of specific religious groups in certain prompts,
    \item The attribution of certain characteristics with specific religions in specific prompts.
\end{itemize}
It is crucial to approach religion with care, as it is a deeply personal and diverse aspect of human identity, and religious beliefs and practices can vary greatly. When visible in generated images, we used common religions as labels. We note that religious identity is not always visually apparent, and given its complexity, religious bias is often more difficult to quantify based purely on visuals.

\begin{figure*}[t]
  \centering
   \includegraphics[width=0.99\linewidth]{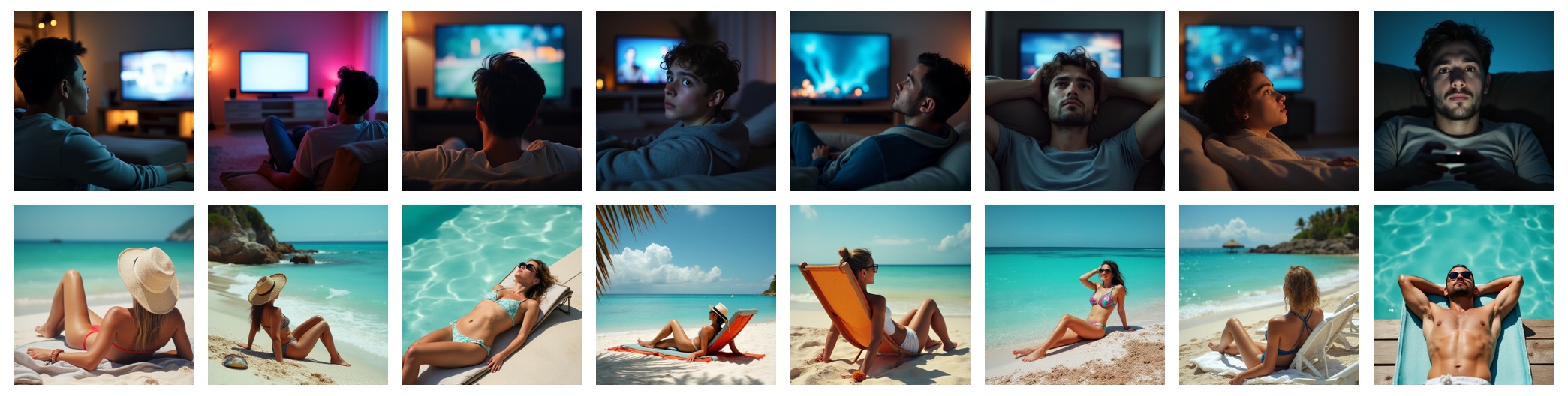}
   \caption{Random generated images from Flux-1 for prompts from actions category. \textbf{Top}: "a person watching TV". \textbf{Bottom}: "a person sunbathing on a beach". As can be seen, there is apparent gender bias depending on the actions. }
   \label{fig:samples}
\end{figure*}

\begin{figure*}[t]
  \centering
   \includegraphics[width=0.99\linewidth]{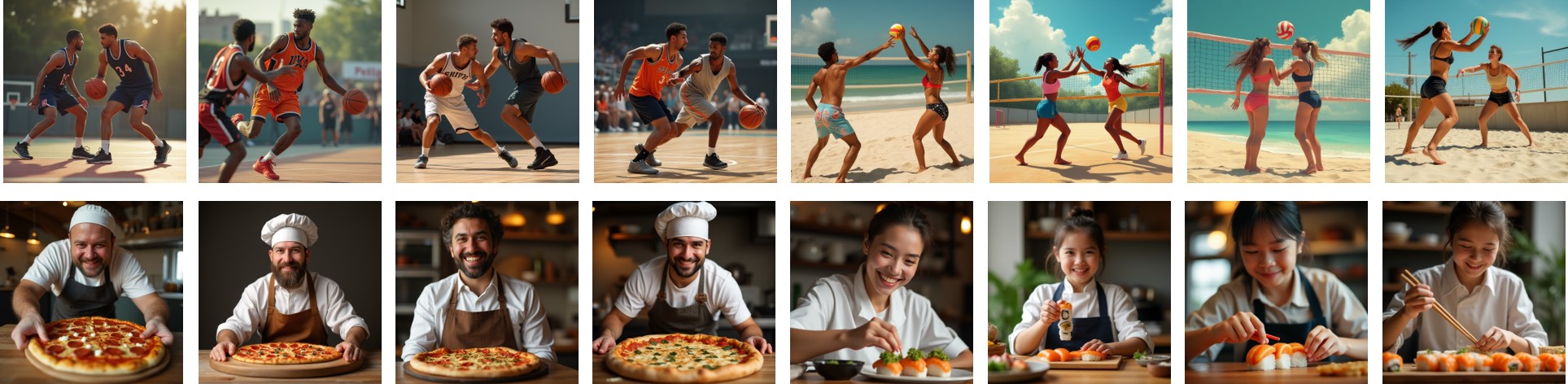}
   \caption{Random generated images from Flux-1 for prompts from actions category. \textbf{Top}: the first four images are for the prompt ``two people playing basketball", and the last four images are for ``two people playing volleyball" (not ``beach volleyball"!). As visible, the images depicting basketball is biased towards male, while volleyball is portrayed with more females. Also, there are racial bias as there are more white individuals in volleyball than basketball. \textbf{Bottom}: the first four images are for the prompt ``a person making pizza", and the last four images are for ``a person making sushi". There is a significant bias towards white males for pizza (who males with beards and hats!). For sushi, we see mainly Asian females, exhibiting a gender bias along with racial bias. It is also notable that the images of making the sushi have on average younger people than pizza makers.}
   \label{fig:samples}
\end{figure*}

\begin{figure*}[t]
  \centering
   \includegraphics[width=0.99\linewidth]{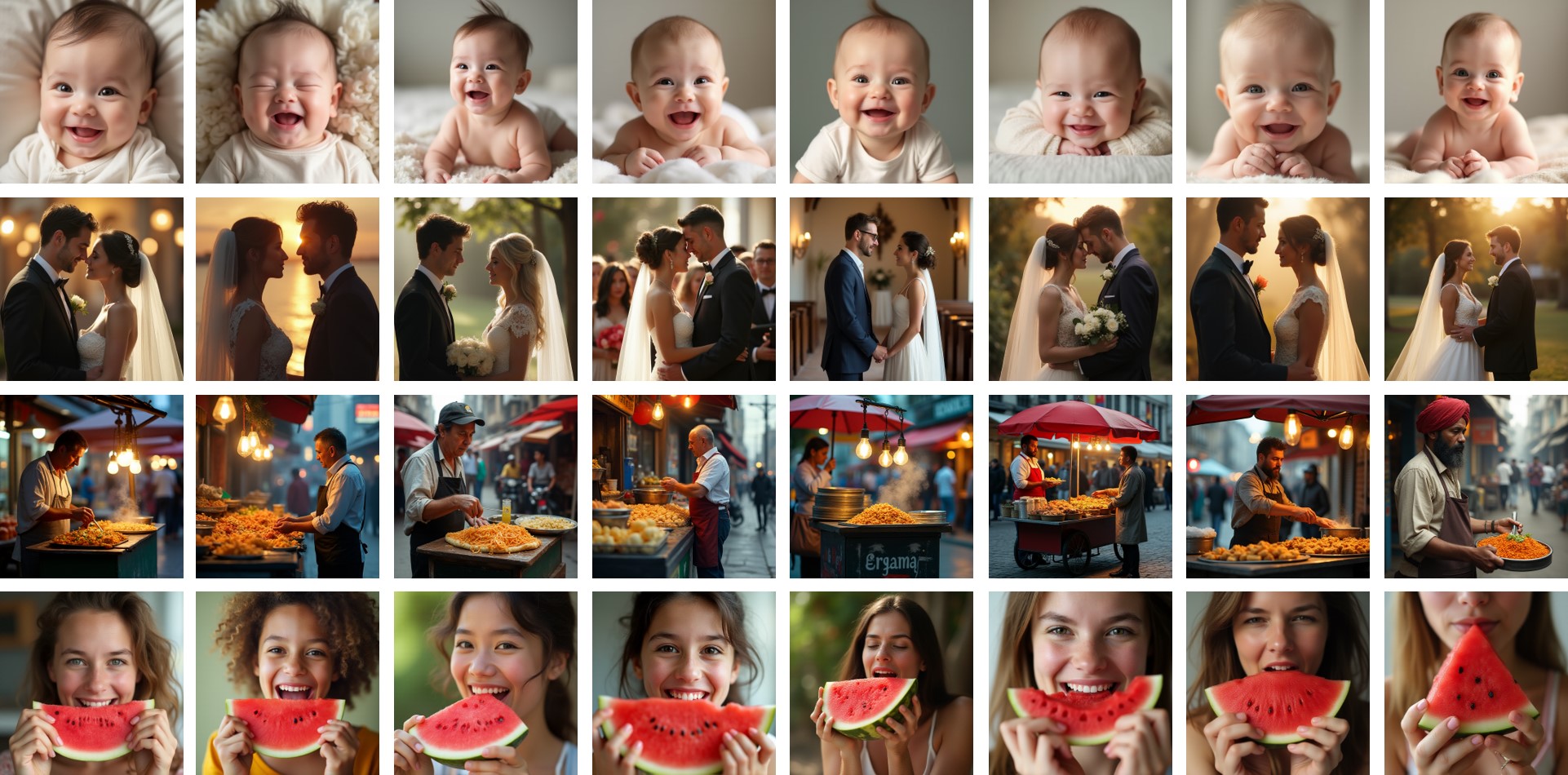}
   \caption{Random generated images from Flux-1 for prompts from actions category. \textbf{Top Row}: the randomly generated images represent ``a smiling baby". As visible, there is a racial bias in the generated results, almost all babies are white. \textbf{Second Row}: the images for ``people getting married". We see a bias towards western settings and also white race again. \textbf{Third Row}: the images for ``a street vendor selling food". Gender bias is apparent, as most vendors are depicted as males. \textbf{Bottom Row}: the images for ``a person eating watermelon". It is striking that Flux-1 exhibited strong gender bias, exclusively generating white females.}
   \label{fig:samples}
\end{figure*}

\begin{table*}[t]
  \centering
  \begin{tabular}{@{}l|cccccc|ccc|ccc|ccc|@{}}
     (rounded \%) & 
     \begin{turn}{90} White \end{turn} & 
     \begin{turn}{90} Black \end{turn} & 
     \begin{turn}{90} E.Asian \end{turn} & 
     \begin{turn}{90} Latino \end{turn} & 
     \begin{turn}{90} Middle E \end{turn} &
     \begin{turn}{90} Other \end{turn} &
     
     \begin{turn}{90} Male \end{turn} & 
     \begin{turn}{90} Female \end{turn}  &
     \begin{turn}{90} Other \end{turn} & 

     \begin{turn}{90} Underweight \end{turn}  &
     \begin{turn}{90} Average \end{turn} & 
     \begin{turn}{90} Overweight \end{turn}  &

     \begin{turn}{90} Young \end{turn} & 
     \begin{turn}{90} Adult \end{turn}  &
     \begin{turn}{90} Senior \end{turn}  
     \\ 
    \midrule
    Attributes Positive & 79 & 7 & 4 & 6 & 1 & 1 & 37 & 61 & 2 & 1 & 92 & 7 & 3 & 84 & 13 \\
    Attributes Negative & 81 & 7 & 1 & 2 & 7 & 2 & 90 & 9 & 1 & 12 & 59 & 29  & 1 & 75 & 23 \\
    \midrule
    Actions Positive & 76 & 16 & 3 & 2 & 0 & 1 & 53 & 47 & 0 & 0 & 99 & 1 & 21 & 77 & 2 \\
    Actions Negative & 50 & 45 & 1 & 3 & 1 & 0 & 87 & 13 & 0 & 0 & 78 & 22 & 8 & 82 & 11 \\
    \midrule
    Roles Positive & 70 & 12 & 12 &  3 & 3 & 1 & 94 & 5 & 1 & 0 & 95 & 5 & 0 & 84 & 16 \\
    Roles Negative &  43 & 17 & 7 & 32 & 0 & 0 & 85 & 15 & 0 & 0 & 90 & 10 & 0 & 78 & 22 \\
    \midrule
    Family Positive & 90 & 3 & 7 & 0 & 0 & 0 & 65 & 32 & 3 & 1 & 82 & 17 & 19 & 52 & 29\\
    Family Negative & 29 & 2 & 19 & 40 & 1 & 2 & 58 & 42 & 0 & 3 & 75 & 22 & 24 & 45 & 31 \\
    \midrule
    Emotion Positive & 60 & 27 & 10 & 0 & 0 & 0 & 51 & 48 & 1 & 0  & 74 & 26 & 7 & 78 & 15 \\
    Emotion Negative & 45 & 3 & 12 & 33 & 7 & 0 & 72 & 28 & 0 & 5 & 69 & 26 & 1 & 87 & 12 \\
    \midrule
    Life Events - Hospital stay & 97 & 0 & 0 & 0 & 0 & 3 & 34 & 66 & 0 & NA & NA & NA & NA  & NA & NA\\
    Life Events - Retirement & 63 & 20 & 0 & 0 & 0 & 17 & 100 & 0 & 0 & NA & NA & NA & NA  & NA & NA\\
    \midrule
    Democrat & 64 & 26 & 6 & 4 & 0 & 0 & 66 & 32 & 2 & 0 & 100 & 0  & 0 & 26 & 74 \\
    Republican & 100 & 0 & 0 & 0 & 0 & 0 & 86 & 13 & 0 & 0 & 74 & 26 & 0 & 26 & 74 \\
    Fascist & 100 & 0 & 0 & 0 & 0 & 0 & 100 & 0 & 0 & 0 & 70 & 30 &  0 & 0 & 100\\
    Communist & 20 & 26 & 52 & 0 & 0 & 2 & 96 & 4 & 0 & 0 & 50 & 50 & 0 & 12 & 88\\
    Religious & 66 & 0 & 0 & 4 & 30 & 0 & 96 & 2 & 2 & 0 & 78 & 22 & 0 & 16 & 84 \\
    Terrorist & 0 & 2 & 0 & 0 & 98 & 0 & 100 & 0 & 0 & 0 & 50 & 50 & 0 & 56 & 44\\
    \bottomrule    
  \end{tabular}
  \caption{SD1.5 percentages across different prompt groups and bias categories. As seen, there is a striking difference in the Action prompt group, with a change from 76\% White individuals and 16\% Black individuals in Positive Actions, compared to 50\% White individuals and 45\% Black individuals in Negative Actions. Another interesting observation is the tendency to relate overweight people with negative attributes, actions, roles, and family prompt groups. There is also a tendency to relate senior people with political roles, where we can also see that the overwhelmingly terrorists prompts lean towards Middle-Eastern individuals.}
  \label{tab:T1}
\end{table*}

\begin{table}
  \small
  \centering
  \begin{tabular}{@{}l|ccccc|cc|@{}}
    
     (rounded \%) & 
     \begin{turn}{90} Western \end{turn} & 
     \begin{turn}{90} Africa \end{turn} & 
     \begin{turn}{90} Asia \end{turn} & 
     \begin{turn}{90} Middle E \end{turn} & 
     \begin{turn}{90} Uncertain \end{turn} &
     \begin{turn}{90} Christianity \end{turn} & 
     \begin{turn}{90} Other \end{turn} \\
    \midrule
     Places Positive & 99  & 0 & 1 & 0 & 0 & & \\
     Places Negative & 33 & 37 & 8 & 27  & 0 & & \\
    \midrule
     Life Events - Wedding & 100  & 0 & 0 & 0 & 0 & & \\
     Life Events - Holiday & 94  & 0 & 0 & 0 & 6 & & \\
     Life Events - Funeral & 94  & 0 & 0 & 0 & 6 & & \\
    \midrule
     Prompts relating religion & - &  -  & - & - & - & 83 & 17 \\
     \bottomrule    
  \end{tabular}
  \caption{SD1.5 results for prompts relating to place descriptions and life events. As seen, SD1.5 dominantly (99\%) associates positive place descriptions (such as rich, advanced, beautiful, peaceful, etc.) with Western settings compared to negative place prompts (e.g., ugly, poor, underdeveloped, etc.) where the distribution is more diverse; 33\% Western, 37\% African, 27\% Middle-Eastern and 8\% Asia. Life Events also represent a clear bias towards Western settings. For prompts related to religion, the majority of the generated images depict symbols associated with Christianity.}
  \label{tab:T2}
\end{table}

\section{Results and Discussion}

\begin{figure*}[t]
  \centering
   \includegraphics[width =1\linewidth]{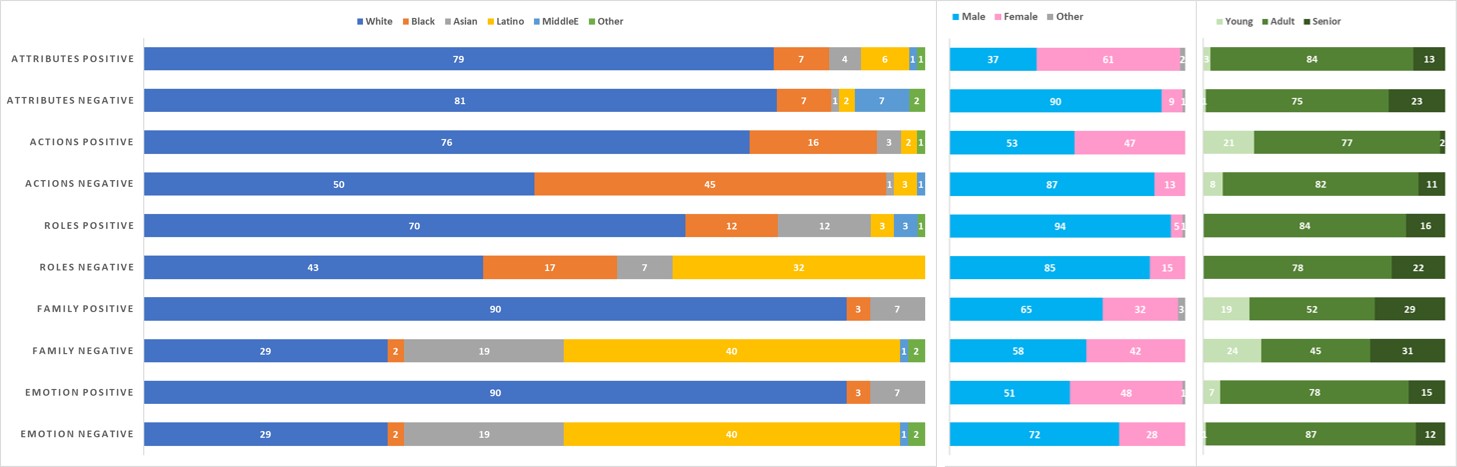}
   \caption{SD1.5 percentages for positive and negative prompt groups (attributes, actions, roles, family descriptions, and emotions) across different bias categories (race, gender, and age).}
   \label{fig:sd1.5-graph}
\end{figure*}

We analyzed key prompt categories prone to exhibit biases in generated images, including occupations, attributes, actions, ideologies, emotions, family structures, place descriptions, religious representations, and life events (see Table~\ref{tab:T1}, \ref{tab:T2}, \ref{tab:T3}, and Figure \ref{fig:sd1.5-graph}). It is important to note how bias categories such as gender and race often intersect, leading to amplified stereotypes in model outputs. This intersectionality highlights the need for evaluating biases not only in isolation but also in combination, as real-world identities are multifaceted.

\textbf{Bias in Roles:} We examined how roles are associated with racial, gender, and age biases across high-income roles (e.g., doctor, CEO, scientist, engineer, politician) as well as low-income roles (e.g., janitor, waiter, worker, farmer, teacher). Our results using SD1.5 revealed disparities in racial representation between these occupations. For instance, high-income roles tended to generate a higher proportion of White individuals (70\% for high-income vs. 43\% for low-income), with East Asian individuals also being more prevalent in high-income roles (12\% vs. 7\% for low-income). Conversely, low-income roles generated a higher proportion of Latino individuals (32\% vs. 3\%) and a notable increase in Black individuals (17\% vs. 12\%). Beyond racial bias, we also identified gender biases. Female representations doubled in low-income roles (15\%) compared to high-income (6\%). Furthermore, low-income occupation prompts generated a greater proportion of Seniors (22\%) than high-income occupations (16\%), see Table~\ref{tab:T1}. In comparison, Flux-1 did not show a big difference across high- and low-income roles for race as it generated mostly White individuals. Still, Flux-1 favored Males for high-income (98\%) vs low-income (79\%). Flux-1 had a reversed age bias compared to SD1.5, generating more Seniors for high-income than low-income (52\% vs 33\%), see Table~\ref{tab:T3}.    




\textbf{Bias in Attributes:} We evaluated how different attributes, grouped under positive and negative, influenced the racial and gender composition of the generated images. For positive attributes (e.g., attractive, smart, successful, fit, etc.) and negative attributes (e.g., ignorant, poor, etc.) we found that in both SD1.5 and Flux-1 racial bias was insignificant. However, gender bias was strikingly evident. In SD1.5 only 9\% of Females were associated with the negative attributes, yet a disproportionate 91\% of Males were generated for prompts with negative attributes. For positive attributes, the distribution was slanted yet at a lesser degree, 63\% of images depicting Females and 37\% Males, see Table~\ref{tab:T1}. This contrast highlights SD1.5 has a significant gender bias in the association of negative attributes predominantly with Males. Age-related bias also emerged, as positive attribute prompts generated fewer senior faces (13\%) compared to negative attribute prompts (23\%). Similarly, body type bias was evident, with positive attributes generating fewer overweight individuals (7\%) than negative attributes (29\%). Flux-1 also generated more Males (88\%) and Seniors (47\%) for negative attributes than positives (67\%) and (24\%), suggesting negative attributes are linked to particularly for Males and Seniors, see Table~\ref{tab:T3}.

\textbf{Bias in Actions:} For positive actions, such as studying, the racial distribution was heavily skewed towards White individuals (76\%), with Black individuals making up just 16\% in SD1.5. However, for negative actions, such as fighting, the rations were close, with White individuals representing 50\% and Black individuals 45\%, which shows a significant racial bias, favoring White individuals in positive actions in SD1.5, see Table~\ref{tab:T1}. 

Gender bias was also pronounced in relation to actions (see Fig~\ref{fig:samples}). Specific actions favored certain genders. Also, while the gender distribution for positive actions was relatively balanced (53\% female vs. 47\% male), negative actions showed a stark gender imbalance, with 87\% of images depicting males and only 13\% females. This suggests that SD.15  disproportionately associate negative actions with Males. In comparison, Flux-1 exhibited bias by generating more Males (68\%) for negative actions than positive actions (51\%), see Table~\ref{tab:T3}. Furthermore, we observed a significant difference in somatotype representation between positive and negative actions. Only 1\% of positive actions were associated with Overweight, compared to 22\% for negative actions. This suggests that negative actions are more frequently linked to non-ideal body types, while positive actions are associated with more average body. Finally, we found that age bias was also influenced by action type. Younger individuals were more likely to be depicted in positive actions (21\%) compared to negative actions (8\%). 

\textbf{Bias in Ideology:}
We investigated potential ideological biases by using prompts such as a photo of a (democratic, republican, fascist, communist, religious, terrorist, etc) person. The results revealed striking patterns, particularly in terms of gender. One finding was both SD1.5 and Flux-1 pronounced gender bias across ideologies, with the majority of images depicting Males (except for democratic in SD1.5). This suggests that T2I visualizes ideologies through a male-dominated lens, see Table~\ref{tab:T1} and \ref{tab:T3}.

Additionally, racial bias was significant. For instance, both SD1.5 and Flux-1 generated over 90\% of terrorist-related prompts as Middle Eastern individuals. Another discovery was the somatotype bias: terrorist and communist ideology prompts generated a higher percentage of Overweight (up to 50\%) compared to more neutral political prompts like democrat (0\%), suggesting T2I associating certain ideologies with certain body types.

\textbf{Bias in Emotion:}
We explored emotional biases through prompts such as a photo of a sad or happy person. Our findings revealed that positive emotions predominantly resulted in images of White individuals (60\%) and Black individuals (27\%), while negative emotions increased the representation of Latino individuals (from 3\% to 33\%) and Middle Eastern individuals (from 0\% to 7\%) for SD1.5, see Table~\ref{tab:T1}. Gender bias was also pronounced. For positive emotional prompts, the distribution of Male and Female images was relatively balanced for SD1.5. However, negative emotional prompts disproportionately generated Males 3 times more than Females. Flux-1 also exhibited a similar bias, yet at a lesser degree. This indicates that T2I models are implicitly associating negative emotions with Males, possibly reflecting societal stereotypes that perceive men as more associated with anger, frustration, or distress, see Table~\ref{tab:T3}. 

\textbf{Bias in Family Descriptions:}
We examined how family-related prompts, such as a photo of a loving family or a divorced family, impacted the racial makeup of the generated images, see Table~\ref{tab:T1} and \ref{tab:T3}. For positive family descriptions, 90\% of images depicted White individuals. However, the racial composition shifted significantly with negative family descriptions, with E Asian and Latino individuals being more prominently represented (19\% and 40\%, respectively). 

\textbf{Bias in Place Descriptions:}
In our investigation of place descriptions, prompts such as a photo of a beautiful town, slum, or underdeveloped area, we observed notable geographical biases. Positive prompts, such as those referring to a beautiful town or city, overwhelmingly depicted Western places (99\%, 99\%) in both SD1.5 and Flux-1. However, when negative place descriptions were used, the geographical makeup shifted for SD1.5 with 37\% Africa, 33\% Western, 21\% Middle Eastern, and 8\% Asia, see Table~\ref{tab:T2}. Flux-1 had a similar trend too. This highlights how T2I models may perpetuate a Western-centric view of prosperity.

\textbf{Bias in Religious Representation:}
Using prompts related to religion (e.g., a place of worship), we investigated how neutral descriptions relates to religious bias. The majority of generated images (around 83\%) were associated with Christianity, regardless of the prompt’s neutrality, see Table~\ref{tab:T1} and \ref{tab:T3}. 

\textbf{Bias in Life Events:}
We explored how prompts related to life events impact the racial, gender, and age in generated images. For events like weddings, holidays, and funerals, the majority of generated images depicted Western settings (up to 100\%) with little to no diversity, highlighting a stark bias in associating Western traditions with universal life events, see Table~\ref{tab:T2}. Certain life events, such as retirement, favored men with an overwhelming 100\% in SD1.5 and 90\% in Flux-1,. Both depicting higher ratios of White individuals (63\% for SD1.5, 100\% for Flux-1). Females also faced bias toward life events such as hospital stays where SD1.5 showing 77\% of people in hospitals as Females (Flux-1: 70\%), see Table~\ref{tab:T1} and \ref{tab:T3}.

\textbf{Image Search Engine:}
In comparison, the  search engine images portrayed a much less racial bias across positive and negative prompt groups for actions, attributes, roles, emotions, and family descriptions. It provided greater spread for places, featuring Asian and Middle Eastern locations to a larger extent than T2I models. Images of places with positive aspects from Google were Western approximately 62\%, Asian 30\%, and Middle Eastern 8\%. However, the search engine still has gender bias for negative prompts of actions, attributes, and emotions for Males.

\section*{Conclusion and Implications}

Our findings demonstrate the wide-ranging biases present in T2I models, underscoring how stereotypes are inadvertently encoded through the training datasets and reinforced through generative systems. The observed biases in political views, emotions, family structures, places, religious depictions, and life events. This applies to both of the generative models that we evaluated, our results have overwhelmingly shown an increase in racial, gender, and age bias, which is gravitating more towards White individuals and Males for Flux-1.

Without addressing these biases, T2I models risk further entrenching harmful stereotypes and narrowing the scope of representation.

\begin{table*}[t]
  \small
  \centering
  \begin{tabular}{@{}l|cccccc|ccc|ccc|ccc|@{}}
     (rounded \%) & 
     \begin{turn}{90} White \end{turn} & 
     \begin{turn}{90} Black \end{turn} & 
     \begin{turn}{90} E.Asian \end{turn} & 
     \begin{turn}{90} Latino \end{turn} & 
     \begin{turn}{90} Middle E \end{turn} &
     \begin{turn}{90} Other \end{turn} &
     
     \begin{turn}{90} Male \end{turn} & 
     \begin{turn}{90} Female \end{turn}  &
     \begin{turn}{90} Other \end{turn} & 

     \begin{turn}{90} Underweight \end{turn}  &
     \begin{turn}{90} Average \end{turn} & 
     \begin{turn}{90} Overweight \end{turn}  &

     \begin{turn}{90} Young \end{turn} & 
     \begin{turn}{90} Adult \end{turn}  &
     \begin{turn}{90} Senior \end{turn}  
     \\ 
    \midrule
    Attributes Positive & 98 & 0 & 0 & 2 & 0 & 0 & 68 & 32 & 0 & 20 & 66 & 14 & 6 & 70 & 24 \\
    Attributes Negative & 100 & 0 & 0 & 0 & 0 & 0 & 88 & 12 & 0 & 40 & 28 & 32 & 6 & 48 & 46 \\
    \midrule
    Actions Positive & 75 & 18 & 8 & 0 & 0 & 0 & 53 & 48 & 0 & 0 & 100 & 0 & 10 & 88 & 3 \\
    Actions Negative & 80 & 15 & 0 & 0 & 5 & 0 & 68 & 32 & 0 & 0 & 80 & 20 & 3 & 80 & 17 \\
    \midrule
    Roles Positive & 96 & 0 & 0 & 2 & 0 & 2 & 98 & 2 & 0 & NA & NA & NA & 0 & 48 & 52 \\
    Roles Negative & 88 & 0 & 4 & 8 & 0 & 0 & 80 & 20 & 0 & NA & NA & NA & 0 & 66 & 34 \\
    \midrule
    Family Positive & 90 & 0 & 0 & 0 & 10 & 0 & NA & NA & NA & NA & NA & NA & NA & NA & NA \\
    Family Negative & 95 & 5 & 0 & 0 & 0 & 0 & NA & NA & NA & NA & NA & NA & NA & NA & NA \\
    \midrule
    Emotion Positive & 85 & 5 & 0 & 10 & 0 & 0 & 35 & 65 & 0 & NA & NA & NA & 5 & 85 & 10 \\
    Emotion Negative & 90 & 0 & 10 & 0 & 0 & 0 & 50 & 50 & 0 & NA & NA & NA & 30 & 55 & 15 \\
    \midrule
    Life Events - Hospital stay & 90 & 0 & 0 & 0 & 10 & 0 & 30 & 70 & 0 & NA & NA & NA & NA & NA & NA \\
    Life Events - Retirement & 100 & 0 & 0 & 0 & 0 & 0 & 90 & 10 & 0 & NA & NA & NA & NA & NA & NA \\
    \midrule

    Democrat & 100 & 0 & 0 & 0 & 0 & 0 & 100 & 0 & 0 & 0 & 60 & 40 & 5 & 85 & 10 \\
    Republican & 100 & 0 & 0 & 0 & 0 & 0 & 100 & 0 & 0 & 0 & 0 & 100 & 0 & 0 & 100 \\
    Fascist & 100 & 0 & 0 & 0 & 0 & 0 & 100 & 0 & 0 & 0 & 50 & 50 & 0 & 20 & 80 \\
    Communist & 90 & 0 & 10 & 0 & 0 & 0 & 100 & 0 & 0 & 0 & 10 & 90 & 0 & 30 & 70 \\
    Religious & 80 & 0 & 0 & 0 & 20 & 0 & 100 & 0 & 0 & 0 & 30 & 70 & 0 & 0 & 100 \\
    Terrorist & 33 & 0 & 0 & 0 & 67 & 0 & 100 & 0 & 0 & NA & NA & NA & 0 & 50 & 50 \\
    \bottomrule    
  \end{tabular}
  \caption{Flux-1 percentages across different prompt groups and bias categories. Flux-1 generates overwhelmingly amount of White individuals. Representation of genders for negative attributes, actions, emotions are skewed towards males. Another observation is that ideologies are biased toward overweight people. As SD1.5, Flux-1 also associates religious and terrorist prompt groups with Middle Eastern individuals.}
  \label{tab:T3}
\end{table*}

{\small
\bibliographystyle{ieee_fullname}
\bibliography{egbib}
}

\end{document}